\documentclass[preprint]{elsarticle}

\usepackage{amsmath}
\usepackage{amsfonts}
\usepackage{graphicx}
\usepackage{booktabs}
\usepackage{subfigure}
\usepackage{multirow}

\journal{Journal of \LaTeX\ Templates}

\begin{document}

\begin{frontmatter}

\title{Reinterpreting CTC Training as Iterative Fitting}

\author[mymainaddress]{Hongzhu Li}

\author[mymainaddress]{Weiqiang Wang\corref{mycorrespondingauthor}}
\cortext[mycorrespondingauthor]{Corresponding author}
\ead{wqwang@ucas.ac.cn}

\address[mymainaddress]{University of Chinese Academy of Sciences, Beijing, China}

\begin{abstract}
The connectionist temporal classification (CTC) enables end-to-end sequence learning by maximizing the
probability of correctly recognizing sequences during training. The outputs of a CTC-trained model tend
to form a series of spikes separated by strongly predicted blanks, know as the spiky problem. To figure
out the reason for it, we reinterpret the CTC training process as an iterative fitting task that is based on
frame-wise cross-entropy loss. It offers us an intuitive way to compare target probabilities with model
outputs for each iteration, and explain how the model outputs gradually turns spiky. Inspired by it, we
put forward two ways to modify the CTC training. The experiments demonstrate that our method can
well solve the spiky problem and moreover, lead to faster convergence over various training settings.
Beside this, the reinterpretation of CTC, as a brand new perspective, may be potentially useful in other
situations. The code is publicly available at https://github.com/hzli-ucas/caffe/tree/ctc.
\end{abstract}


\end{frontmatter}

\section{Introduction}
The connectionist temporal classification (CTC) \cite{Graves2006Connectionist} is a method to solve sequence-to-sequence learning, and is widely used in various sequence labeling tasks, such as speech recognition \cite{Graves2014Towards,Miao2016EESEN,Kim2017Joint}, text recognition \cite{Alex2009A,He2016Reading,Borisyuk2018Rosetta}, dynamic gesture recognition \cite{Molchanov2016Online}, sound event detection \cite{wang2019connectionist}, action labeling \cite{Huang2016Connectionist} and lip reading \cite{Cheng2018Visual}. The basic idea of CTC is to interpret the network outputs, a sequence of probability distributions over different labels, as a probability distribution over all possible label sequences. With an extra $blank$ class, the output at each timestep, or frame, of the sequence indicates either a specific label or no label. The outputs over all timesteps consist a sequence of labels and blanks, named as a $path$. A path is mapped to a label sequence by removing the repeated labels then the blanks in it, in this way a label sequence usually has more than one corresponding path. The CTC training is to maximize the probability of the ground-truth label sequence conditioned on the input, which is calculated by summing up probabilities of all the corresponding paths.

\begin{figure}
  \centering
  \includegraphics[width=0.8\columnwidth]{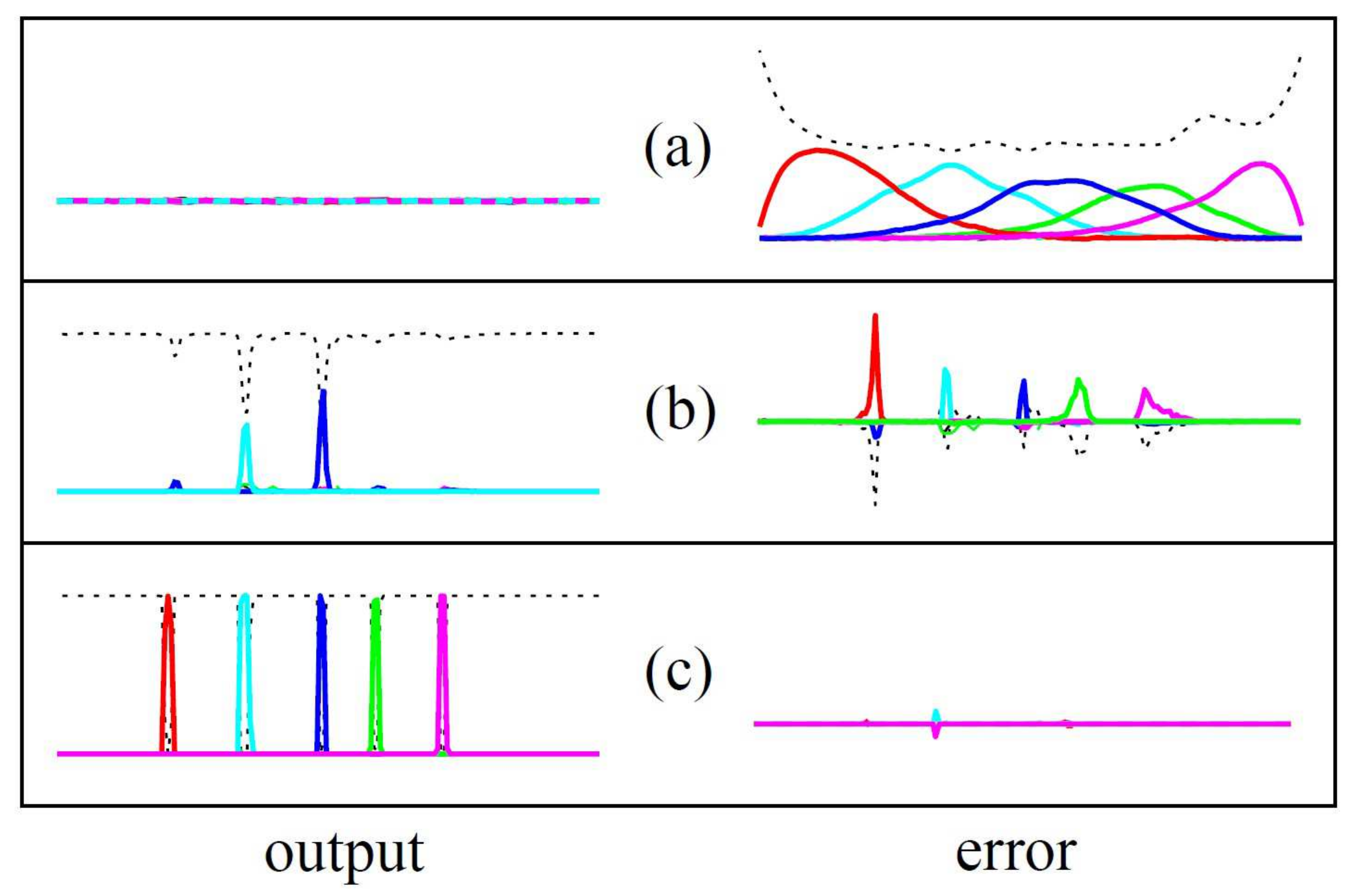}
  \caption{Evolution of the network outputs and CTC error signal during training. Lines with different colors denote different labels, and the dashed line is the $blank$ class. \cite{Graves2006Connectionist}}
  \label{fig:ctc-output}
\end{figure}

Though there's no explicit restriction on the ratio of labels to blanks during training, the outputs of a CTC-trained model tend to form a series of label spikes separated by strongly predicted blanks as in Figure~\ref{fig:ctc-output} \cite{Graves2006Connectionist}. This phenomenon is welcomed in some situation, for example, Wang and Metze \cite{wang2019connectionist} aims to predict the begin/end of each sound event, where a spike signal is wanted. On the other hand, the spiky distribution is not desirable when the model needs to densely predict labels for consecutive frames, and may harm the model generalization for its implying that non-blank labels is predicted in strict condition. According to Liu et al. ​\cite{Liu2018Connectionist}, the CTC training will concentrate on a dominant path once it finds one, and dominant paths are often overwhelmed by blanks as blanks are included in most of the feasible paths, thus leads to the spiky problem. They propose a maximum entropy regularization for CTC (EnCTC), which prevents the entropy of feasible paths from decreasing too fast, to enhance exploration during training and get smoother and wider activation of non-blank labels. Huang et al. ​\cite{Huang2016Connectionist} adopt CTC for action labelling in video, getting a degenerated spiky path, which is dominated by a single non-blank label while blank other labels appears as spikes. They utilize prior knowledge to supervise CTC training, re-weighting paths with frame-to-frame visual similarities and ruling out infeasible paths with a few sparsely annotated frames, to get a reasonable path. In order to speed up training, Sampled CTC \cite{Variani2018Sampled} proposes to sample only one path for optimization in each iteration to speed up training for recurrent neural networks, it reduces the spikiness of CTC as a side-effect. All these works consider the probability of correct labelling as the sum of probability of corresponding paths. This is certainly based on the definition of CTC, but not visually intuitive to be connected with the spiky problem.

We put aside the $path$ perspective and consider each $frame$ as a unit.
Based on a pseudo GT (ground-truth), we replace the original loss function of the CTC, the negative log probability of correct
labelling the sequence, with the sum of frame-wise cross entropy.
Assuming the gradients of two loss functions are equal, the pseudo
GT can be derived from the original CTC update formula. It should
be noted that, for each frame within each sequence, the pseudo
GT is not a fixed value during the training process, and its value
is determined by the ground-truth label sequence and the current
model outputs. Therefore, compared with cross-entropy optimization based on true frame-wise labeling, CTC training can be viewed
as a heuristic iterative frame-wise fitting process. The reinterpretion of CTC does not change the training process, but brings a new
perspective to understand and modify it.

From this perspective, we provide a tool to simulate and display
the iterative training process of CTC, and analyse why CTC training leads to spikes. A very intuitive way is proposed to deal with
the spikes, which is to directly adjust the non-blank label proportion in pseudo GT. Observing that the model’s convergence speed
is mainly affected by key frames, which usually have large error
signals, we focus training on key frames to speed up the convergence process.

The main contributions are summarized as follows: (1) We
reinterpret the CTC objective function as frame-wise cross entropy,
to understand and improve CTC from a new perspective; (2) Modify the pseudo GT to solve the CTC spike problem, where we can
simply use a hyper-parameter to specify the label proportion in
network outputs; (3) Accelerate network convergence by frame
reweighting, and experiments confirm its effectiveness at different
training settings; (4) We provide a visualization tool that simulates
the CTC training process, which can be used to observe the training
process of CTC, as well as peek the effects of further modifications
on CTC to rule out the bad ones in advance.

\section{Method}

\subsection{Connectionist Temporal Classification}\label{sec:ctc}

The CTC \cite{Graves2006Connectionist} is proposed for labeling sequence data within a single network architecture that doesn't need pre-segmentation and post-processing. The basic idea is to interpret the network outputs as a conditional probability distribution over all possible output label sequences. Given this distribution, an objective function can be derived that directly maximises the probabilities of the correct label sequences.

At each timestep, the network outputs a probability distribution over the label set $L'=L\cup\{blank\}$, where $L$ contains all the labels in the task and the extra $blank$ represents `no label'. The activation $y^t_k$ is interpreted as the probability of observing label $k$ of $L'$ at time $t$. Given the length $T$ input sequence $\mathbf{x}$, we get the conditional probability $p(\pi|\mathbf{x})$ of observing a particular $path$ $\pi$ through the lattice of label observations:
\begin{equation}
  p(\pi|\mathbf{x})=\prod_{t=1}^Ty^t_{\pi_t},\forall\pi\in L'^T,
  \label{equ:path-prob}
\end{equation}
where $\pi_t$ is the label observed at time $t$ along path $\pi$, and $L'^T$ is the set of length $T$ paths over $L'$.

Paths are mapped onto label sequences by an operation $\mathcal{B}$ that simply removes the repeated labels then the blanks in a sequence. For a given label sequence $\mathbf{l}\in L^U,{U\leq T}$, more than one $\pi$ corresponds to it, e.g. $\mathcal{B}(aa--ab-)=\mathcal{B}(-a-abb)=aab$, where `$-$' denotes the $blank$. We can evaluate the conditional probability of $\mathbf{l}$ as the sum of probabilities of all the corresponding paths:
\begin{equation}
  p(\mathbf{l|x})=\sum_{\pi\in\mathcal{B}^{-1}(\mathbf{l})}p(\pi |\mathbf{x}).
  \label{equ:seq-prob}
\end{equation}
The calculation seems problematic, for the the number of corresponding paths grows exponetially with $U$. It can be solved with a dynamic-programming algorithm similar to the forward-backward algorithm for HMMs \cite{Rabiner1993A}​, by breaking down the sum over paths corresponding to a labelling $\mathbf{l}$ into an iterative sum over paths corresponding to prefixes of that labelling. To allow for blanks in the output paths, we consider a modified label sequence $\mathbf{l}'\in L′^{2U+1}$, with blanks added at the beginning and the end of $\mathbf{l}$, and between every pair of consecutive labels. In calculating the probabilities of prefixes of $\mathbf{l}'$, transitions between blank and non-blank labels, and between any pair of distinct non-blank labels are allowed. Further calculation details can be found in \cite{Graves2012Connectionist}​. 

The CTC loss function is defined as the negative log probability of correctly labelling the sequence:
\begin{equation}
  {\rm CTC}(\mathbf{l,x})=-{\rm ln}\,p(\mathbf{l|x}).
  \label{equ:ctcLoss}
\end{equation}
During training, to backpropagate the gradient through the output layer, we need the derivatives of the loss function versus the outputs $\{a^t_k|t\in[1,T],k\in L'\}$ before the activation function is applied. For the softmax activation function
\begin{equation}
  y^t_k=\frac{e^{a^t_k}}{\sum_{k'}e^{a^t_{k'}}},
\end{equation}
where $k'$ ranges over $L'$, the derivative with respect to $a^t_k$ is
\begin{equation}
  \frac{\partial{\rm CTC}(\mathbf{l,x})}{\partial a^t_k}=y^t_k-\frac{1}{p(\mathbf{l|x})}\sum_{\substack{\pi\in\mathcal{B}^{-1}(\mathbf{l}):\\ \pi_t=k}}p(\pi|\mathbf{x}),
  \label{equ:ctc-grad}
\end{equation}
where $\sum_{\pi\in\mathcal{B}^{-1}(\mathbf{l}):\\ \pi_t=k}p(\pi|\mathbf{x})$ is the sum of probabilities of all the paths corresponding to $\mathbf{l}$ that go through the label $k$ at time $t$. The Equation (\ref{equ:ctc-grad})​ is an equivalent of Equation (7.34) in ​\cite{Graves2012Connectionist}​ by substituting Equation (7.25) into it.

When the network is used for prediction, the predictions over all timesteps are converted into a label sequence. Since the computational complexity grows exponentially with the length of the path, it is not practical to find the most probable label sequence $\mathbf{\hat{l}}$. There are many approximate alternatives, and the \textbf{best path decoding} is one of the most commonly used methods. It assumes that the most probable output will correspond to $\mathbf{\hat{l}}$:
\begin{equation}
  \begin{aligned}
    \mathbf{\hat{l}}&\approx\mathcal{B}(\pi^*)\\
    {\rm where}\ \pi^* &=\mathop{\arg\max}_{\pi}{p(\pi|\mathbf{x})}.
  \end{aligned}
\end{equation}
It is not guaranteed to find the most probable label sequence, but the solution is good enough in most cases and the computation procedure is trivial.

\subsection{Cross Entropy}
The cross entropy (CE) is used to estimate the distance between two probability distributions. Given ground-truth $\mathbf{y'}$ and network outputs $\mathbf{y}$, the cross entropy loss is defined as
\begin{equation}
  {\rm CE}(\mathbf{y',y})=-\sum_k y'_k{\rm ln}(y_k),
  \label{equ:ceDef}
\end{equation}
where $k$ ranges over all the classes, $y_k$ and $y'_k$ are the model's estimated and ground-truth probabilities for class $k$ respectively.
Let $\{a_k\}$ be the model's outputs before the softmax activation function is applied, the loss function derivative with respect to $a_k$ can be found by
\begin{equation}
  \frac{\partial{\rm CE}(\mathbf{y',y})}{\partial a_k}=y_k-y'_k.
  \label{equ:ceGrad}
\end{equation}

The cross-entropy objective function is usually used for classification problems, where $\mathbf{y'}$ is a one-hot distribution. In our opinion, since the definition of cross entropy does not restrict $\mathbf{y'}$ to one-hot distribution, this objective function is also suitable for fitting a general probability distribution (not necessarily one-hot) to update $\mathbf{y}$ toward it.

\subsection{Reinterpretion of CTC}
\label{sec:ctcCE}

Given an input sequence $\mathbf{x}$ and its ground-truth label sequence $\mathbf{l}$, the network outputs probability distributions $\mathbf{Y}=\{y^t_k|t\in[1,T],k\in L'\}$ over the $T$ timesteps of the sequence. In the original CTC formula Equation (\ref{equ:path-prob}), only the input $\mathbf{x}$ id considered as a conditional parameter. But the caluculation is actually performed on $\mathbf{Y}$, which is not only determined by the input, but also related to the current network parameter $\mathbf{W}$, which should not be ignored. According to us, $p(\mathbf{l|x})$ should be $p(\mathbf{l|x,W})$ or $p(\mathbf{l|Y})$. For more accurate expression in the following, we will replace $\mathbf{x}$ in the formula with $\mathbf{Y}$ when necessary.

We define $\mathbf{y}_t=\{y^t_k|k\in L'\}$ as the predicted probability distribution for the sample of timestep $t$, and assume there is a corresponding ground-truth probability distribution $\mathbf{y'}_t=\{y'^t_k|k\in L'\}$. The cross entropy should be equivalent to the original CTC loss, meaning that they behave the same during gradient back-propagation, so there is
\begin{equation}
  \frac{\partial{\rm CTC}(\mathbf{l,Y})}{\partial a^{t}_{k}}=\frac{\partial\sum_{t'}{\rm CE}(\mathbf{y'}_{t'},\mathbf{y}_{t'})}{\partial a^{t}_{k}}
  \label{equ:ctcCE}
\end{equation}
A feasible solution for $\mathbf{y'}_t$ can be found by following the conditions below:
\begin{equation}
  \begin{cases}
    y'^t_k=\frac{1}{p(\mathbf{l|Y})}\sum_{\substack{\pi\in\mathcal{B}^{-1}(\mathbf{l}):\\ \pi_t=k}}p(\pi|\mathbf{Y}),\\
    \frac{\partial y'^t_k}{\partial y^{t'}_{k'}}=0,\forall t,t'\in[1,T],k,k'\in L'.
  \end{cases}
  \label{equ:ctcGT}
\end{equation}
We can get the derivative of $\sum_t{\rm CE}(\mathbf{y'}_t,\mathbf{y}_t)$ versus $a^t_k$
\begin{equation}
  \frac{\partial\sum_{t'}{\rm CE}(\mathbf{y'}_{t'},\mathbf{y}_{t'})}{\partial a^t_k}=y^t_k-y'^t_k,
  \label{equ:ctc-grad2}
\end{equation}
which equals to the CTC loss function derivative as in Equation (\ref{equ:ctc-grad}). See the appendix for proof process.

It seems unreasonable that the derivative of $y'^t_k$ versus $y^{t'}_{k'}$ equals to zero, when $y'^t_k$ depends on $\mathbf{Y}$. We argue that the pseudo GT (ground-truth) $\mathbf{Y}'$ is calculated on $\mathbf{Y}$, but used as a constant during each iteration. It sounds sophistical, but is acceptable if the CTC training process is understood as an iterative fitting task as follows. For each iteration, given the current output $\mathbf{Y}$ and label sequence $\mathbf{l}$, the pseudo GT is calculated in a heuristic way as
\begin{equation}
  y'^t_k=\frac{1}{p(\mathbf{l|Y})}\sum_{\substack{\pi\in\mathcal{B}^{-1}(\mathbf{l}):\\ \pi_t=k}}p(\pi|\mathbf{Y})=\frac{p(\mathbf{l},\pi_t=k|\mathbf{Y})}{p(\mathbf{l|Y})}=p(\pi_t=k|\mathbf{l,Y}),
\end{equation}
It is a posterior probability, that frame t takes category k, conditioned on that the network output $\mathbf{Y}$ is correctly mapped to the ground-truth label sequence $\mathbf{l}$. Let $\mathbf{Y'}$ be the target of the current iteration, and fit the network output $\mathbf{Y}$ to it based on the cross-entropy loss.

\begin{figure}
  \centering
  \includegraphics[width=0.8\columnwidth]{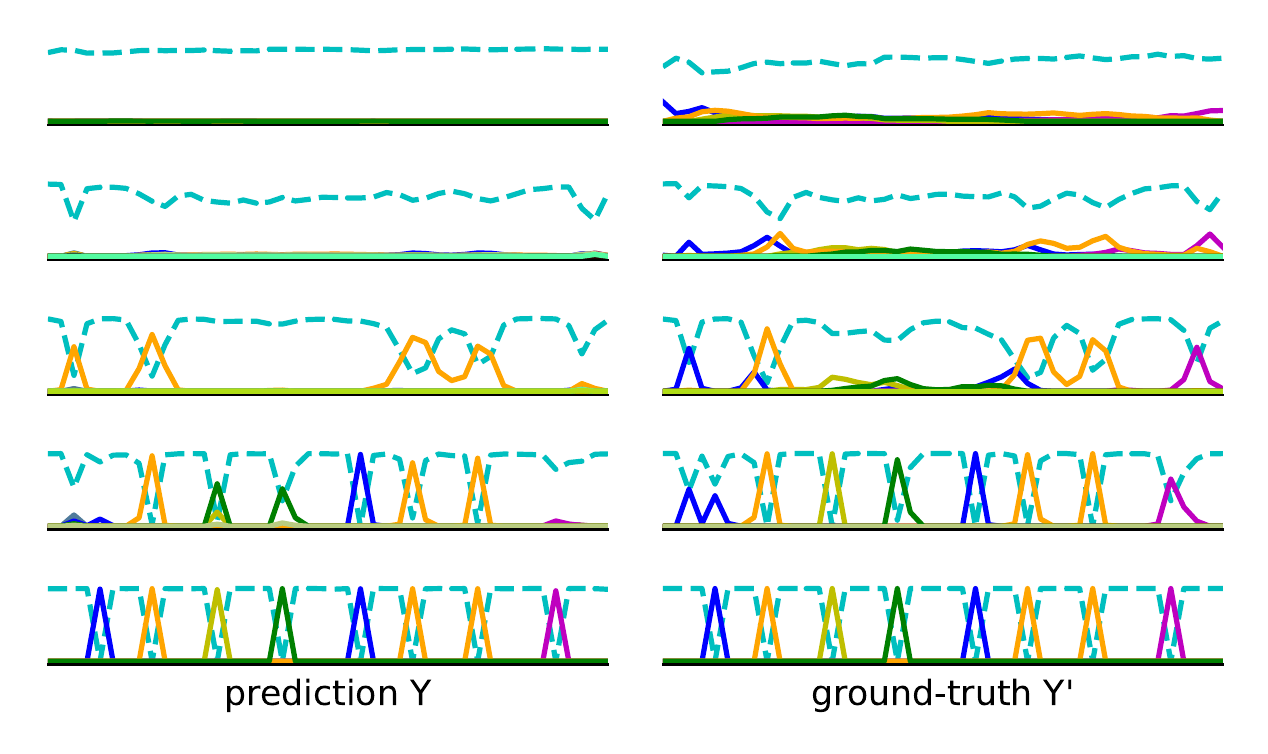}
  \caption{Examples of predicted probability distributions $\mathbf{Y}$ and the corresponding ground-truths $\mathbf{Y'}$. This is the same sequence during different iterations. Lines with different colors denote different labels, and the dashed line indicates $blank$.}
  \label{fig:predGT}
\end{figure}

Some examples of $\mathbf{Y}$ and the corresponding $\mathbf{Y'}$ are illustrated in Figure~\ref{fig:predGT} for an intuitive perception. It is easy to find that the pseudo GT is actually the sum of the network output and the error signal in ​Figure~\ref{fig:ctc-output}. Note that the reinterpretion in this section does not really change CTC, but aims to provide a different perspective to understand CTC.

\subsection{Analysis of Spiky Problem}

In outputs of CTC-trained models, the non-blank label activations usually appear in spiky shapes, and most frames are occupied by high-confidence blanks. To find out the reason for it, we simulate and display the training process of CTC with a simple tool, where we use a randomly initialized matrix as network output, update it according to CTC, and draw the output probability distributions and pseudo GT for each iteration. Observing the training process presented in ​Fig. 4​, we consider the spiky problem as a result of the following two aspects.

\begin{figure}
  \centering
  \includegraphics[width=1.0\columnwidth]{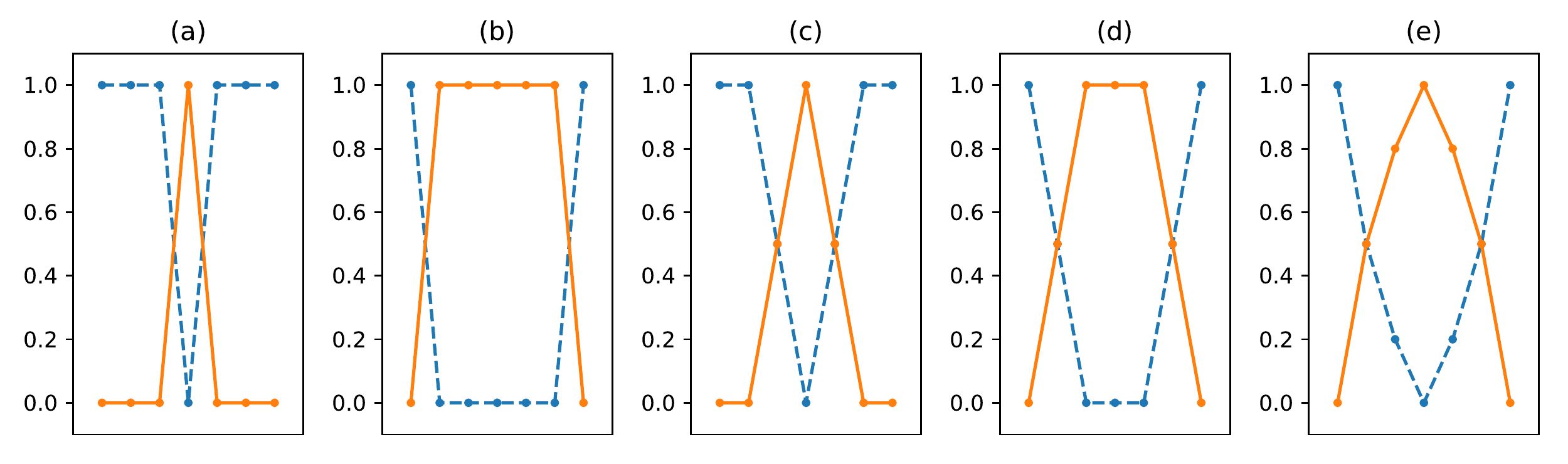}
  \caption{Examples of probability distribution along time-axis. The solid line indicates character `a', and the dash line indicates $blank$. The dots on lines indicate the frame segmentation.}
  \label{fig:prob}
\end{figure}

\begin{figure}
  \centering
  \includegraphics[width=1.0\columnwidth]{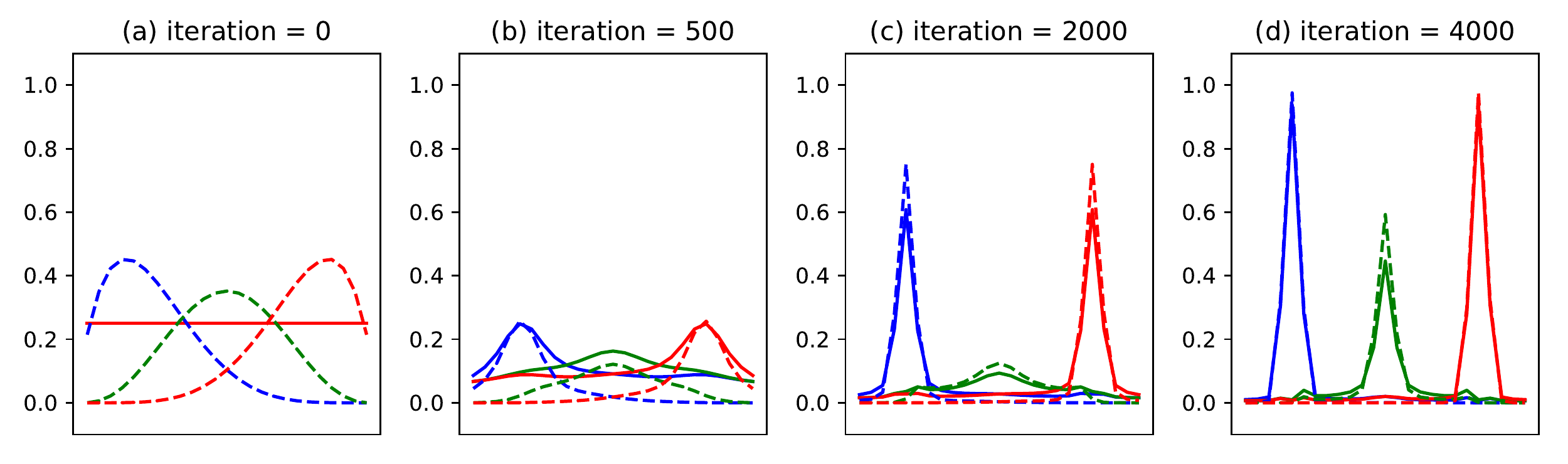}
  \caption{The convergence process simulation of CTC training. Lines with different colors present the activations of different non-blank labels, where solid lines indicate the model outputs and dash lines indicate the pseudo GT.}
  \label{fig:sim}
\end{figure}

On one hand, CTC takes sequence probability as the optimization objective, so it prefers outputs that are correctly mapped to the label sequence with high probabilities. We can get an intuitive perception from the probability distributions given in Figure~\ref{fig:prob}, setting $a$ as the label sequence. Sub figure (a) and (b) respectively obtains a path, $---a---$ and $-aaaaa-$, with 100\% probability. Both paths are mapped to $a$ after removing of repeated labels then blanks, so the probability of the correct label sequence is 100\%. Sub figure (c) and (d) both consist of multiple paths. For example, (c) contains path $--aaa--$, $--aa---$, $---aa--$, and $---a---$ with 25\% probability each. All the four paths are mapped to $a$, so the probability of correct labelling is still 100\%. As for sub figure (e), the probability of the correct label sequence is only 81\%, because it contains paths that are mapped to wrong label sequences, for example $\mathcal{B}(-a-a---)=aa$. Comparing the shapes of these distributions, we find that the smoother the shape, the greater probability of wrong paths. Therefore, the target probability distribution always tend to be steeper than the current output, as can be viewed in Figure~\ref{fig:sim}.

On the other hand, the outputs of randomly initialized networks usually have uniform probability distributions, e.g. Figure~\ref{fig:ctc-output}(a), and it takes a while before the outputs become steep enough. During this period, the model convergence would be biased towards the dominant category, and we call this phase the \textbf{suppression phase}. In CTC training, $blank$ is usually the dominant class for blanks consist more than half of the modified label sequence $\mathbf{l}'$ as mentioned in Section~\ref{sec:ctc}. The suppression by $blank$ can be observed in ​Figure~\ref{fig:sim}(b), which is very obvious on the middle label. When the output is steep enough, it is occupied by high-confidence blanks then, the probability of the label gradually increases to form a spike like ​Figure~\ref{fig:prob}(a) or (c) instead of ​​Figure~\ref{fig:prob}(b) or (d). We call this process the \textbf{peaking phase}, and illustrate it in ​Figure~\ref{fig:sim}​(c) and (d).

\subsection{Setting the Non-blank Proportion}

Observing the convergence process of CTC training, we find that the proportion of labels decreases to zero during the suppression phase but cannot increase too much after entering the peaking phase. Since the spikes lead to the low proportion of non-blank category, maintaining the proportion at a reasonable value seems useful to solve the spiky problem. We use a hyperparameter $\alpha$ to specify the non-blank proportion in the sequence, and make sure the proportion of non-blanks be $\alpha$ and the proportion of blanks be $1-\alpha$ during the entire convergence process. The most direct way is to modify the pseudo GT by scaling down categories that exceed the desired proportions and scaling up categories that do not reach the proportion.

At first we calculate the total amount of each category in the sequence, 
\begin{equation}
  V_k=\sum_t y'^t_k,
\end{equation}
and then calculate the proportion that this category should occupy
\begin{equation}
  N_k=\sum_i \mathbb{I}(\mathbf{l}_i=k)\text{ where }k\neq blank,
\end{equation}
which is assumed to be the number of occurrences in the label sequence. $\mathbb{I}(\cdot)$ equals 1 when the condition is true, otherwise 0. The proportion of $blank$ should be the length of the tag sequence, which is elaborated in Section~\ref{sec:ctc}. Based on hyperparameter $\alpha$, we scale the values for different class of pseudo GT as
\begin{equation}
  y'^t_k\leftarrow
  \begin{cases}
    y'^t_k (1-\alpha)(\sum_{k'\neq blank}N_{k'})/V_k & \text{if}\ k=blank\\
    y'^t_k \alpha N_k/V_k & \text{otherwise}
  \end{cases}
\end{equation}
For each frame, the sum of the values over different categories is normalized to one to obtain a valid probability distribution 
\begin{equation}
  y'^t_k\leftarrow\frac{y'^t_k}{\sum_{k'}y'^t_{k'}}.
\end{equation}

\begin{figure}
  \centering
  \includegraphics[width=1.0\columnwidth]{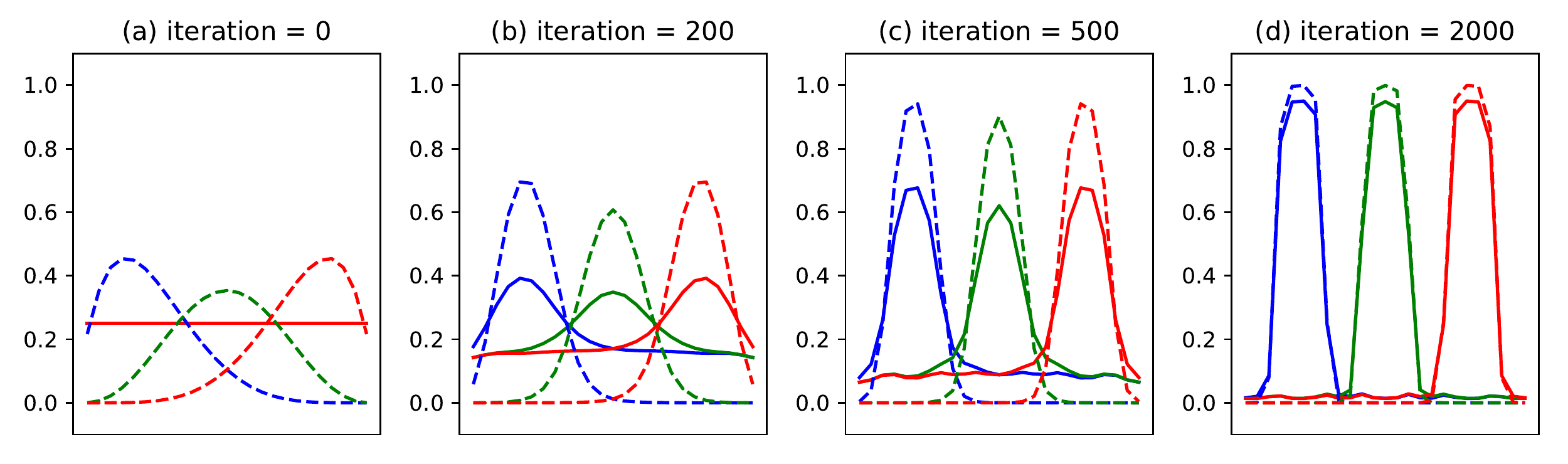}
  \caption{The convergence process simulation of the modified CTC training that maintains the non-blank ratio of pseudo GT to be $\alpha=0.5$. Lines with different colors present the activations of different non-blank labels, where solid lines indicate the model outputs and dash lines indicate the pseudo GT.}
  \label{fig:sim-alpha}
\end{figure}

This is a highly simplified rescaling strategy, where different non-blank labels are assumed to have the same proportion, the rescaling is performed within each batch instead of sequence, and the non-blank proportion after normalization will only be an approximation of $\alpha$. We carry out a simulation experiment for this method, and obtain the result in Figure~\ref{fig:sim-alpha}, which is is satisfactory for solving the spiky problem. Besides, the method also shows an effect of speeding up training, for ideal outputs are obtained around iteration 2000, way ahead compared with ​Fig. 4​. It may have something to do with the absence of suppression phase.

\subsection{Focusing on Key Frames}

We find that different frames in a sequence are of different importance during the CTC convergence. The most important ones are those that prevent the label shapes being more steeper, i.e. the hillside frames during suppression phase and the peak frames during peaking phase, named as $key\ frames$. The key frames usually have a large probability increase in a certain category, which is $blank$ in the suppression phase, and non-blank in the peaking phase. Based on the ideas of key frames, we reweight frames within each sequence to focus the training on key frames, and use $\gamma$ as the hyperparameter to adjust weighting degree. Following GHM \cite{Li2019Gradient}, which performs reweighting direclty on gradient rather than on loss, we reweight the gradient as
\begin{equation}
  \frac{\partial{\rm CTC}_\gamma(\mathbf{l,Y})}{\partial a^t_k}=w_t^\gamma(y^t_k-y'^t_k),
\end{equation}
where the frame weight is calculated by
\begin{equation}
  w_t=\max_k(y'^t_k-y^t_k).
\end{equation}
Then within each sequence, the gradient is divided by the weights sum for normalization. 
It is easy to notice that when the value of $\gamma$ is 0, this modification makes no difference.

\begin{figure}
  \centering
  \includegraphics[width=1.0\columnwidth]{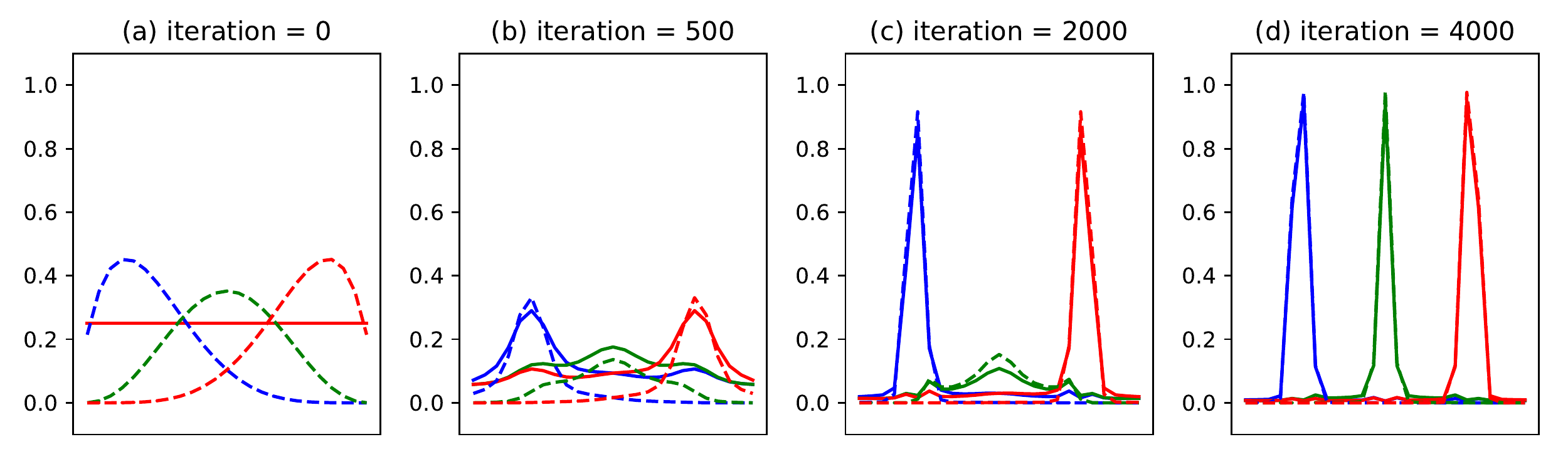}
  \caption{The convergence process simulation of the modified CTC training that focuses on key frames with $\gamma=1$. Lines with different colors present the activations of different non-blank labels, where solid lines indicate the model outputs and dash lines indicate the pseudo GT.}
  \label{fig:sim-gamma}
\end{figure}

The simulation experiment is shown in Figure~\ref{fig:sim-gamma}, where this method lead to faster convergence than basic CTC training as in Figure~\ref{fig:sim}.

\section{Experiments}

To evaluate the effects of our proposed method, we compare them with the CTC training according to the output distributions, convergence process, generalization and final accuracy of the models.
For all the experiments, the accuracy refers to sequence accuracy, i.e. the percentage of testing images correctly recognized.

\subsection{Datasets}
\label{sec:dataset}

For all the following experiments, we use the synthetic dataset (Synth90k) released by Jaderberg et al. ​ \cite{Jaderberg2014Synthetic} as the training data. The dataset consists of 8 million word images and their corresponding ground-truth words. All the images are generated by a synthetic data engine using a 90k word dictionary, and are of different sizes. This is the training set used in all of our experiments, unless otherwise emphasized.

There are four popular benchmarks for scene text recognition used for model performance evaluation, namely IIIT5k-word (IIIT5k), Street View Text (SVT), ICDAR 2003 (IC03) and ICDAR 2013 (IC13). \textbf{IIIT5k} \cite{IIIT5k} contains 3,000 cropped word images collected from the Internet. \textbf{SVT} \cite{SVT} contains 647 word images cropped from 249 street-view images that are collected from Google Street View. \textbf{IC03} \cite{IC03} contains 251 scene images, we discard words that either contain non-alphanumeric characters or have less than three characters, and get 860 cropped word images. \textbf{IC13} \cite{IC13} contains 1,095 word images in total, we discard words that contain non-alphanumeric characters, and get 1,015 word images with ground-truths.

To further evaluate the model generalization performance, we conduct experiments on a small dataset \textbf{Synth5k} \cite{Liu2018Connectionist}, which consists of 5k training data and 5k testing data randomly sampled from Synth90k.

\subsection{Implementation Details}
\label{sec:detail}
We use CRNN \cite{Shi2017An} as our baseline model, and follow their training settings to train the network on Synth90k. During training, all word images are scaled to size $32\times100$ ignoring their aspect-ratios.

We implement the network architecture within the Caffe \cite{Jia2014Caffe} framework, with custom implementation for the input and loss layer. To obtain comprehensive results, we adopt different gradient decent optimization algorithms and learning rates in our experiments, and the details will be presented in each section. The batch-size is set to 100 in all the experiments.

For all the experiments, we get the recognition results by the lexicon-free best path decoding \cite{Graves2006Connectionist}.

\subsection{Complexity Analysis}
We propose two different ways to modify the CTC training, and compare the algorithm complexities of them to the original CTC training. Let $L$ denote the batch-size, $T$ denote the length of input sequence, $U$ denote the length of the label sequence $\mathbf{l}$, and $C$ denote the output class number $|L'|$. The back-propagated gradient is based on $y^t_k$ and $y'^t_k$, which are calculated as follows. First, a softmax activation is applied to get the normalized network outputs $\{y^t_k\}$, whose time complexity is $O(NTC)$ for the cpu implementation and $O(C)$ for the parallel gpu implementation. Then a dynamic-programming algorithm is performed to calculate $\{y'^t_k\}$, whose time complexity is $O(NTU)$ for cpu and $O(TU)$ for gpu implementations. For basic CTC training, there is the final subtraction, whose time complexity is $O(NTC)$ for cpu and $O(1)$ for gpu. Meanwhile, the proposed methods need additional operations. We obtain the time complexity of CTC by summing up the above terms, and list the time complexity of the additional operations for each method in Table~\ref{tab:complexity}. It's obvious that the additional operations do not change the time complexity, so the modified training will not take much more time than CTC training. This is also validated by experiments, where the changes of training time are negligible.

\begin{table}
\caption{The additional time complexity and training time for the modification on CTC, compared with the CTC training. Trn. Time(GPU) denotes training time spent for 100,000 iterations on a single Geforce Titan GPU.}
\label{tab:complexity}
\vskip 0.15in
\begin{center}
\begin{small}
\begin{sc}
\begin{tabular}{cccc}
\toprule
Method & Complexity-cpu & Complexity-gpu & Trn. Time(GPU) \\
\midrule
$\rm CTC$ & $O(NTC)+O(NTU)$ & $O(C)+O(TU)+O(N)$ & 197min \\
\midrule
$\rm CTC\ with\ \alpha$ & $O(NTC)+O(TU)$ & $O(C)+O(T)+O(U)+O(N)$ & 199min \\
$\rm CTC\ with\ \gamma$ & $O(NTC)$ & $O(C)$ & 194min \\
\bottomrule
\end{tabular}
\end{sc}
\end{small}
\end{center}
\vskip -0.1in
\end{table}

The space complexity of CTC is $O(NTC)$. Since the algorithm makes the most of the original space, the $\gamma$ modification needs no additional space, and the additional space complexity of the $\alpha$ modification is only $O(C)$, which doesn't change the original space complexity.

In one word, the proposed methods have the same time and space complexity as the CTC training.

\subsection{Adjustment of Non-Blank Proportion}

\begin{figure}
  \centering
  \includegraphics[width=0.6\columnwidth]{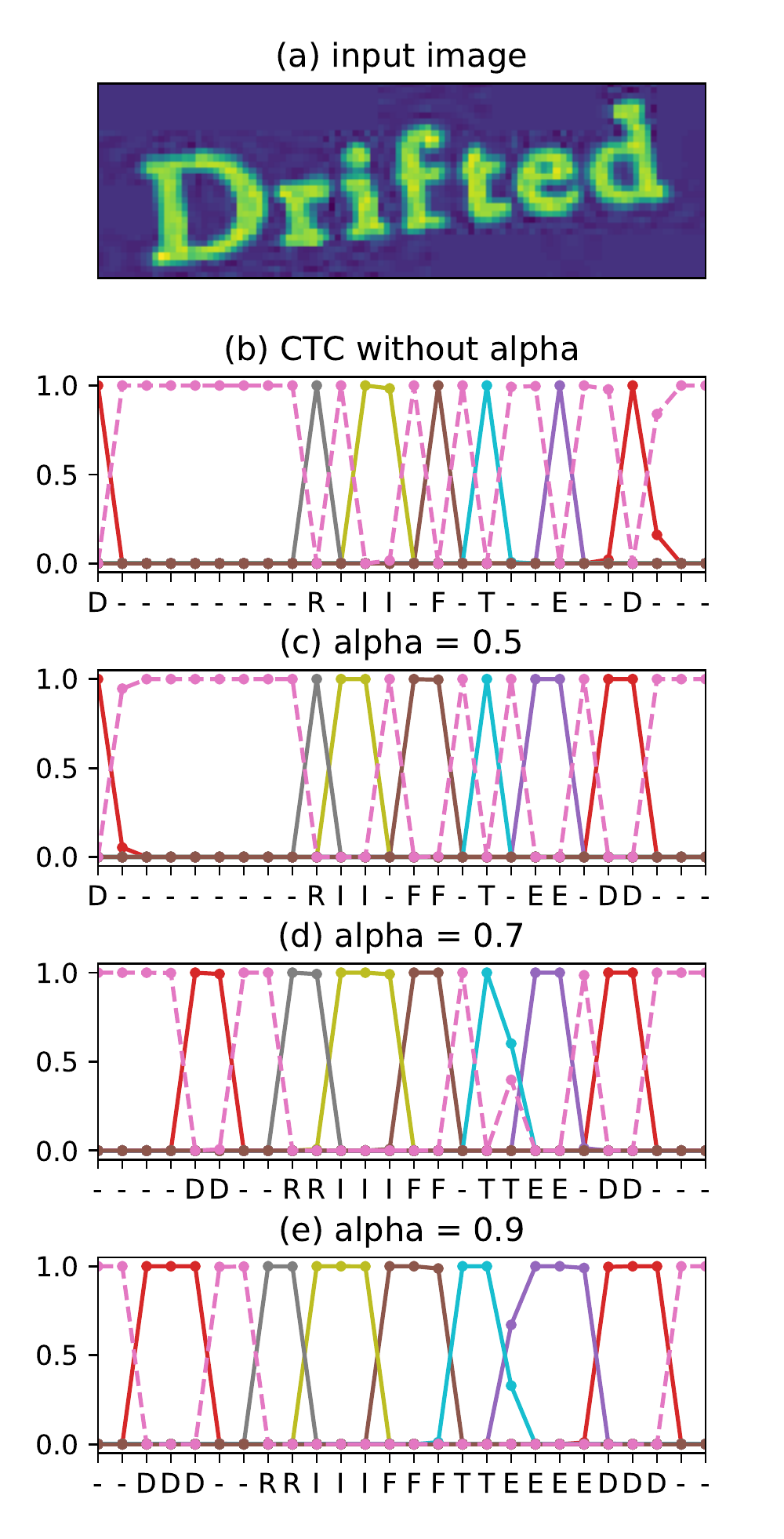}
  \caption{Illustration of the prediction results of an example image. Lines with different colors denote different labels, and the dashed line indicates $blank$. The labels under the horizontal axis indicate the class with the maximum probability at each timestep.}
  \label{fig:output}
\end{figure}

When use $\alpha$ to adjust the label proportion in network outputs, we obtain very ideal effects in the simulation experiment. We conduct experiments on real-world data and get the results that are highly consistent with the simulation. The output of models trained with different $\alpha$ are illustrated in Figure~\ref{fig:output}.

Compared with EnCTC \cite{Liu2018Connectionist} fixing the spiky problem with maximum entropy regularization, our method has one more advantage: we can specify the ratio of the labels according to actual needs, and freely adjust the width of the labels on the time axis.

\subsection{Facilitation of Convergence}
\label{sec:facilitate}

We conduct control experiments to check if the proposed method is effective for speeding up the training. In order to comprehensively validate the effectiveness, we consider three different optimization algorithms, including SGD, Adam, and AdaDelta. For SGD and AdaDelta, we give three control groups at different learning rates, where larger or smaller learning rate has a negative effect on the convergence speed. Therefore, we believe that the comparison experiments are representative of various training settings and the results are convincing. For optimization method Adam, CTC-trained model cannot stably converge at a higher order learning rate 0.01, so we present two sets of controls. For experiments in this section, the learning rate is fixed during training. Every 50k iterations, we test the models on all four test sets, and record the accuracies in Table~\ref{tab:sgd},~\ref{tab:adam}, and~\ref{tab:adadelta}.

\begin{table}
\caption{The accuracy on all testing images from IIIT5k, SVT, IC03, IC13. The models are optimized by SGD with momentum 0.9.}
\label{tab:sgd}
\vskip 0.15in
\begin{center}
\begin{small}
\begin{sc}
  \subtable[learning rate: 0.01]{
      \begin{tabular}{c|cccc}
        \toprule
        iter. & CTC & $\alpha=0.5$ & $\gamma=1$ & $\alpha=0.5,\gamma=1$\\
        \midrule
        50k & 66.8 & 70.6 & 71.5 & 72.0\\
        100k & 73.7 & 75.0 & 75.9 & 75.4\\
        150k & 74.8 & 77.4 & 74.3 & 78.9\\
        200k & 74.9 & 77.1 & 78.1 & 80.0\\
        250k & 76.0 & 76.6 & 76.9 & 79.7\\
        \bottomrule
      \end{tabular}
  }
  \subtable[learning rate: 0.001]{
      \begin{tabular}{c|cccc}
        \toprule
        iter. & CTC & $\alpha=0.5$ & $\gamma=1$ & $\alpha=0.5,\gamma=1$\\
        \midrule
        50k & 69.9 & 72.0 & 74.3 & 74.9\\
        100k & 72.1 & 72.2 & 78.2 & 79.8\\
        150k & 75.4 & 76.7 & 79.0 & 81.0\\
        200k & 77.7 & 77.8 & 79.5 & 81.3\\
        250k & 77.3 & 78.3 & 80.0 & 82.5\\
        \bottomrule
      \end{tabular}
  }
  \subtable[learning rate: 0.0001]{
      \begin{tabular}{c|cccc}
        \toprule
        iter. & CTC & $\alpha=0.5$ & $\gamma=1$ & $\alpha=0.5,\gamma=1$\\
        \midrule
        50k & 67.2 & 68.1 & 69.8 & 70.8\\
        100k & 70.5 & 71.7 & 72.5 & 75.5\\
        150k & 71.9 & 73.5 & 75.6 & 78.2\\
        200k & 72.9 & 73.8 & 76.0 & 80.1\\
        250k & 72.1 & 74.6 & 78.4 & 79.6\\
        \bottomrule
      \end{tabular}
  }
\end{sc}
\end{small}
\end{center}
\vskip -0.1in
\end{table}

\begin{table}
\caption{The accuracy on all testing images from IIIT5k, SVT, IC03, IC13. The models are optimized by Adam.}
\label{tab:adam}
\vskip 0.15in
\begin{center}
\begin{small}
\begin{sc}
  \subtable[learning rate: 0.001]{
      \begin{tabular}{c|cccc}
        \toprule
        iter. & CTC & $\alpha=0.5$ & $\gamma=1$ & $\alpha=0.5,\gamma=1$\\
        \midrule
        50k & 74.1 & 77.0 & 77.9 & 78.3\\
        100k & 79.4 & 77.4 & 78.5 & 80.9\\
        150k & 79.1 & 78.4 & 80.2 & 81.7\\
        200k & 78.6 & 80.0 & 81.2 & 81.7\\
        250k & 80.1 & 80.8 & 81.5 & 82.2\\
        \bottomrule
      \end{tabular}
  }
  \subtable[learning rate: 0.0001]{
      \begin{tabular}{c|cccc}
        \toprule
        iter. & CTC & $\alpha=0.5$ & $\gamma=1$ & $\alpha=0.5,\gamma=1$\\
        \midrule
        50k & 71.2 & 74.4 & 76.1 & 76.2\\
        100k & 73.3 & 77.3 & 79.5 & 79.3\\
        150k & 77.4 & 77.2 & 81.8 & 80.8\\
        200k & 78.8 & 79.2 & 82.1 & 81.0\\
        250k & 78.4 & 81.0 & 81.8 & 82.9\\
        \bottomrule
      \end{tabular}
  }
\end{sc}
\end{small}
\end{center}
\vskip -0.1in
\end{table}

\begin{table}
\caption{The accuracy on all testing images from IIIT5k, SVT, IC03, IC13. The models are optimized by AdaDelta.}
\label{tab:adadelta}
\vskip 0.15in
\begin{center}
\begin{small}
\begin{sc}
  \subtable[learning rate: 10]{
      \begin{tabular}{c|cccc}
        \toprule
        iter. & CTC & $\alpha=0.5$ & $\gamma=1$ & $\alpha=0.5,\gamma=1$\\
        \midrule
        50k & 68.3 & 71.9 & 68.2 & 68.8\\
        100k & 70.0 & 69.7 & 67.2 & 69.3\\
        150k & 69.8 & 70.2 & 69.8 & 67.5\\
        200k & 68.0 & 68.6 & 66.2 & 68.2\\
        250k & 68.2 & 63.8 & 67.7 & 66.0\\
        \bottomrule
      \end{tabular}
  }
  \subtable[learning rate: 1]{
      \begin{tabular}{c|cccc}
        \toprule
        iter. & CTC & $\alpha=0.5$ & $\gamma=1$ & $\alpha=0.5,\gamma=1$\\
        \midrule
        50k & 74.7 & 78.7 & 78.0 & 79.0\\
        100k & 77.1 & 80.8 & 79.8 & 81.3\\
        150k & 80.3 & 80.3 & 79.6 & 81.9\\
        200k & 81.5 & 79.9 & 81.5 & 81.4\\
        250k & 80.6 & 80.9 & 81.6 & 81.6\\
        \bottomrule
      \end{tabular}
  }
  \subtable[learning rate: 0.1]{
      \begin{tabular}{c|cccc}
        \toprule
        iter. & CTC & $\alpha=0.5$ & $\gamma=1$ & $\alpha=0.5,\gamma=1$\\
        \midrule
        50k & 69.3 & 71.9 & 70.0 & 71.8\\
        100k & 73.1 & 74.4 & 74.7 & 75.7\\
        150k & 74.4 & 76.8 & 75.1 & 76.7\\
        200k & 74.3 & 78.9 & 77.5 & 79.5\\
        250k & 76.6 & 77.0 & 77.3 & 79.2\\
        \bottomrule
      \end{tabular}
  }
\end{sc}
\end{small}
\end{center}
\vskip -0.1in
\end{table}

The results show that at most training settings, $\alpha$ and $\gamma$ accelerate the convergence. The extent of acceleration is related to the optimization algorithm and the learning rate. The convergence-facilitation is the most obvious with the basic optimization algorithm SGD, and the proposed methods may cause instability when the learning rate is too large. The acceleration effect usually brings a significant increase in accuracy during early training, but this advantage may gradually diminish as the training continues and model converges.

All in all, comprehensive control experiments demonstrate that our proposed method do speed up convergence, which can be adopted at different training parameters to obtain further acceleration effects.

\subsection{Evaluation on Generalization}
\label{sec:gene}

We follow the practice of EnCTC \cite{Liu2018Connectionist} to evaluate the generalization of models. They train models for 150 epochs on the training data of Synth5k, and compare the generalization of the models with the accuracy on its validation data. They increase the accuracy from 38\% of CTC to 47\% with their proposed method.

However, we believe that their models suffer from under-fitting based on our experimental results. After training our CTC model for 150 epochs, we get similar accuracy as them, which is 41\% in the upper left corner of Table~\ref{tab:gene1}. But continuing to train for another 150 epochs, the accuracy of CTC reaches 50\%, exceeding their best result. So we did not compare the results with them, but still got inspiration from the comparison experiments.

Table~\ref{tab:gene1} shows the accuracies of models after 150 epochs training. At this time, due to insufficient training, the models suffer from underfitting. Our modifications with $\alpha$ and $\gamma$ facilitate the convergence process, thus lead to higher accuracies. Table~\ref{tab:gene2} shows the results after 300 epochs. At this time, CTC model has converged to the optimal solution, but the proposed method could not reach a stable convergence. The same effect can be found in some optimization algorithms , for example Adam, which also accelerates training but cannot reach the optimal solution due to the modification on magnitude and direction of the gradient. We get the results in Table~\ref{tab:gene3} by training the models from Table~\ref{tab:gene1} with \emph{basic CTC} for amother 150 epochs. It can be seen that `first modified CTC then naive CTC' strategy obtains better results than using basic CTC only. We believe that this is due to the proposed methods' heading to better solution space in the early training period, thus getting better generalization.

In addition, experiments on the Synth5k provide us with some experience in choosing hyperparameter values. For example, $\gamma=1$ benifits the convergence process in the previous section, but instead hurts the model performance in this experiment. This shows that even for the same task, the optimal value of hyperparameters may be different for different scales of data, and should chosen carefully. If you cannot be sure, a smaller $\gamma$ for example 0.5 is always safer, may be not very effective but at least no drawbacks.

\begin{table}
\caption{The results of models trained with different hyperparameters on Synth5k. All the models are optimized by AdaDelta with fixed learning rate 1. (a) The models are trained with corresponding hyperparameters for 150 epoch. (b) Trained with different hyperparameters for 150 epoch on the basis of models in (a). (c) Trained with basic CTC for 150 epoch on the basis of models in (a).}
\label{tab:gene}
\vskip 0.15in
\begin{center}
\begin{small}
\begin{sc}
  \subtable[]{
      \label{tab:gene1}
      \begin{tabular}{c|cccc}
        \toprule
        ~ & without $\alpha$ & $\alpha=0.5$ & $\alpha=0.7$ & $\alpha=0.9$ \\
        \midrule
        without $\gamma$ & 41.0 & 43.9 & 43.9 & 42.2\\
        $\gamma=0.5$ & 45.1 & 45.9 & 42.6 & 44.7\\
        $\gamma=1$ & 35.1 & \textbf{47.1} & 45.2 & 38.6\\
        \bottomrule
      \end{tabular}
  }
  \subtable[]{
      \label{tab:gene2}
      \begin{tabular}{c|cccc}
        \toprule
        ~ & without $\alpha$ & $\alpha=0.5$ & $\alpha=0.7$ & $\alpha=0.9$ \\
        \midrule
        without $\gamma$ & \textbf{49.3} & 44.7 & 47.6 & 44.1\\
        $\gamma=0.5$ & 46.9 & 45.8 & 46.3 & 44.8\\
        $\gamma=1$ & 38.7 & 47.4 & 46.8 & 45.4\\
        \bottomrule
      \end{tabular}
  }
  \subtable[]{
      \label{tab:gene3}
      \begin{tabular}{c|cccc}
        \toprule
        ~ & without $\alpha$ & $\alpha=0.5$ & $\alpha=0.7$ & $\alpha=0.9$ \\
        \midrule
        without $\gamma$ & 49.3 & 50.3 & 50.5 & 49.6\\
        $\gamma=0.5$ & 50.8 & 51.6 & \textbf{52.1} & 50.5\\
        $\gamma=1$ & 42.1 & 51.9 & \textbf{52.1} & 50.8\\
        \bottomrule
      \end{tabular}
  }
\end{sc}
\end{small}
\end{center}
\vskip -0.1in
\end{table}

\subsection{Comparison on Accuracy}

Convolutional recurrent neural network (CRNN) \cite{Shi2017An} is one of the most popular methods in CTC-based text recognition, and it is used as the baseline of our work. We compare our methods with CRNN and EnCTC \cite{Liu2018Connectionist} that use the same network structure. Shi et al. ​\cite{Shi2017An} adopt basic CTC to train the model for 250k, and Liu et al. \cite{Liu2018Connectionist} use improved loss function to train each model for about 1200k iterations. We use AdaDelta as optimization algorithm, set initial learning rate to 1, decrease it by a factor of 0.1 at iteration 200k, and end training after 300k iterations. One model is trained with the proposed method, and the other is trained with basic CTC as a control. The accuracies on test sets are presented in Table~\ref{tab:compWithCRNN}. It is reasonable to obtain similar accuracies to CRNN with the model trained with basic CTC, since both models are trained with the same objective function for about the same iterations. The proposed method with hyperparameter $\alpha=0.5,\gamma=1$ achieves a slightly better performance, which is comparable to EnCTC meanwhile spending much less time on training.

\begin{table}
\caption{The results of models with CRNN architecture.}
\label{tab:compWithCRNN}
\vskip 0.15in
\begin{center}
\begin{sc}
\begin{tabular}{ccccc}
\toprule
Method & IIIT5k & SVT & IC03 & IC13\\
\midrule
CRNN \cite{Shi2017An} & 78.2 & 80.8 & 89.4 & 86.7\\
EnCTC \cite{Liu2018Connectionist} & 82.6 & 81.5 & 90.8 & 90.0\\
EsCTC \cite{Liu2018Connectionist} & 81.7 & 81.5 & 92.6 & 87.4\\
EnEsCTC \cite{Liu2018Connectionist} & 82.0 & 80.6 & 92.0 & 90.6\\
\midrule
CTC & 78.8 & 80.5 & 89.2 & 88.4\\
$\alpha=0.5,\gamma=1$ & 81.1 & 82.2 & 91.2 & 87.7\\
\bottomrule
\end{tabular}
\end{sc}
\end{center}
\vskip -0.1in
\end{table}

It has been discussed in section~\ref{sec:facilitate}, as the model converges, the improvement on accuracy brought by accelerated convergence may no longer be obvious. According to section~\ref{sec:gene}, when the model got near the optimal solution, the modified training may bring risks or benefits, depending on the specific usage. When the scale of the training set changes, the effective values of $\alpha$ and $\gamma$ are different, and we need to conduct more experiments to find them. Tuning parameters on such a large-scale training set takes a lot of efforts, and for now we have not found an ``optimal hyperparameter'' that is effective in all situations, which we think as a problem worth further research.

\section{Conclusion}

In this paper, we utilize frame-wise cross-entropy as the loss function, and reinterpret CTC training as a heuristic algorithm, which minimizes its loss through iterative fitting for frame-wise probability distribution. From this perspective, we modify the CTC training in two ways: (1) Modify the target probability to be fitted, which is called pseudo GT here. (2) Reweight the frames within each sequence. Experimental results show that the proposed method can well solve the spiky problem of CTC and facilitate model convergence under different training settings. We also evaluate the effects of our methods on generalization and accuracy of models, but more efforts are needed to find the optimal hyperparameter values.

We provide a tool that simulates and visualizes the training process of CTC, on which we can perform modification and peek its effects on training. For our proposed methods, the simulation results and practical experimental results are quite similar. Though the situation is more complicated in practical training, the simulation experiments can be used as a reference to exclude some useless modifications in advance, which saves us lots of time.

In addition to the two proposed ways, there are more possibilities in modifying CTC from the iterative-fitting perspective. For example, ECTC\cite{Huang2016Connectionist} use sparsely annotated frames to exclude invalid paths in CTC training, which we can perform in a much easier way, by overwriting the model outputs with annotated frames before calculating the pseudo GT. Feng et al \cite{Feng2019Focal}​ combine focal loss \cite{Lin2017Focal}​ with CTC, making the model attend to difficult sequences within each batch, to handle unbalanced datasets. Instead we propose to focus training on difficult frames, which can further deal with imbalance that happens within each sequence. All these are based on the reinterpretation of CTC training, which we believe is potentially useful in more situations.

\section{Acknowledgement}

This work is supported by National Key R\&D Program of China
under contract No. 2017YFB1002203, NSFC projects under Grant
61976201, NSFC Key Projects of International (Regional) Cooperation and Exchanges under Grant 61860206004, and Ningbo
2025 Key Project of Science and Technology Innovation with No.
2018B10071.

\bibliography{reference}

\clearpage
\appendix

\section{Cross Entropy Loss for CTC}

In the paper, we define a pseudo ground-truth $\mathbf{y'}_t=\{y'^t_k|k\in L'\}$, where
\begin{equation}
  \begin{cases}
    y'^t_k=\frac{1}{p(\mathbf{l|Y})}\sum_{\substack{\pi\in\mathcal{B}^{-1}(\mathbf{l}):\\ \pi_t=k}}p(\pi|\mathbf{Y}),\\
    \frac{\partial y'^t_k}{\partial y^{t'}_{k'}}=0,\forall t,t'\in[1,T],k,k'\in L',
  \end{cases}
  \label{apdequ:ctcGT}
\end{equation}
to substitute the CTC loss with the sum of cross entropy losses. To this end, we need to prove that $\mathbf{y'}_t$ is a feasible solution of
\begin{equation}
  \frac{\partial{\rm CTC}(\mathbf{l,Y})}{\partial a^{t}_{k}}=\frac{\partial\sum_{t'}{\rm CE}(\mathbf{y'}_{t'},\mathbf{y}_{t'})}{\partial a^{t}_{k}}
  \label{apdequ:ctcCE}
\end{equation}
It means given the definition of $\mathbf{y'}_t$, Equ. (\ref{apdequ:ctcCE}) holds.

Having $\{a^t_k|t\in[1,T],k\in L'\}$ denote the unnormalized network outputs, we normalize them with the softmax activation,
\begin{equation}
  y^t_k=softmax(a^t_k)=\frac{e^{a^t_k}}{\sum_{k'}e^{a^t_{k'}}}.
\end{equation}
It's easy to know
\begin{equation}
  \frac{\partial y^{t'}_{k'}}{\partial a^t_k}=
  \begin{cases}
    0 & \text{if}\ t'\ne t\\
    y^t_k(1-y^t_k) & \text{if}\ t'=t,k'=k\\
    -y^t_k y^{t}_{k'} & \text{if}\ t'=t,k'\ne k.
  \end{cases}
\end{equation}
The derivation of cross entropy formatted CTC versus $y^t_k$ can be calculated as
\begin{equation}
\begin{aligned}
  & \frac{\partial{\rm CTC}(\mathbf{l,Y})}{\partial y^t_k} = \frac{\partial\sum_{t'}{\rm CE}(\mathbf{y'}_{t'},\mathbf{y}_{t'})}{\partial y^t_k}\\
  =& -\frac{\partial\sum_{t',k'} y'^{t'}_{k'}{\rm ln}(y^{t'}_{k'})}{\partial y^t_k}\\
  =& -\frac{\partial y'^t_k{\rm ln}(y^t_k)}{\partial y^t_k}\\
  =& -y'^t_k\frac{\partial{\rm ln}(y^t_k)}{\partial y^t_k}+{\rm ln}(y^t_k)\frac{\partial y'^t_k}{\partial y^t_k}\\
  =& -\frac{y'^t_k}{y^t_k},
\end{aligned}
\end{equation}
and its derivation with respect to $a^t_k$ can be calculated as
\begin{equation}
\begin{aligned}
  & \frac{\partial{\rm CTC}(\mathbf{l,Y})}{\partial a^t_k} = \frac{\partial\sum_{t'}{\rm CE}(\mathbf{y'}_{t'},\mathbf{y}_{t'})}{\partial a^t_k}\\
  =& \sum_{t'',k'}\frac{\partial\sum_{t'}{\rm CE}(\mathbf{y'}_{t'},\mathbf{y}_{t''})}{\partial y^{t'}_{k'}}\frac{\partial y^{t''}_{k'}}{\partial a^t_k}\\
  =& \sum_{k'}\frac{\partial\sum_{t'}{\rm CE}(\mathbf{y'}_{t'},\mathbf{y}_{t'})}{\partial y^t_{k'}}\frac{\partial y^t_{k'}}{\partial a^t_k}\\
  =& \frac{\partial\sum_{t'}{\rm CE}(\mathbf{y'}_{t'},\mathbf{y}_{t'})}{\partial y^t_k}\frac{\partial y^t_k}{\partial a^t_k}+\sum_{k'\ne k}\frac{\partial\sum_{t'}{\rm CE}(\mathbf{y'}_{t'},\mathbf{y}_{t'})}{\partial y^t_{k'}}\frac{\partial y^t_{k'}}{\partial a^t_k}\\
  =& (-\frac{y'^t_k}{y^t_k})y^t_k(1-y^t_k)+\sum_{k'\ne k}(-\frac{y'^t_{k'}}{y^t_{k'}})(-y^t_k y^t_{k'})\\
  =& y'^t_k y^t_k-y'^t_k+\sum_{k'\ne k}y'^t_{k'} y^t_k\\
  =& y^t_k\sum_{k'}y'^t_{k'}-y'^t_k\\
  =& y^t_k\frac{1}{p(\mathbf{l|Y})}\sum_{\pi\in\mathcal{B}^{-1}(\mathbf{l})}p(\pi|\mathbf{x})-y'^t_k\\
  =& y^t_k-y'^t_k.
\end{aligned}
\end{equation}
It is equal to the derivative of CTC given in the paper, so Equ.(\ref{apdequ:ctcCE}) holds.

\end{document}